# AIM: Acoustic Inertial Measurement for Indoor Drone Localization and Tracking

Yimiao Sun[1], Weiguo Wang[1], Luca Mottola[2], Ruijin Wang[3], Yuan He[1*]

[1]Tsinghua University, [2]Politecnico di Milano, Italy and RI.SE Sweden
[3]University of Electronic Science and Technology of China
{sym21,wwg18}@mails.tsinghua.edu.cn, luca.mottola@polimi.it
ruijinwang@uestc.edu.cn, heyuan@mail.tsinghua.edu.cn

## ABSTRACT

We present Acoustic Inertial Measurement (AIM), a one-of-a-kind technique for indoor drone localization and tracking. Indoor drone localization and tracking are arguably a crucial, yet unsolved challenge: in GPS-denied environments, existing approaches enjoy limited applicability, especially in Non-Line of Sight (NLoS), require extensive environment instrumentation, or demand considerable hardware/software changes on drones. In contrast, AIM exploits the acoustic characteristics of the drones to estimate their location and derive their motion, *even in NLoS* settings. We tame location estimation errors using a dedicated Kalman filter and the Interquartile Range rule (IQR). We implement AIM using an off-the-shelf microphone array and evaluate its performance with a commercial drone under varied settings. Results indicate that the mean localization error of AIM is 46% lower than commercial UWB-based systems in complex indoor scenarios, where state-of-the-art infrared systems would not even work because of NLoS settings. We further demonstrate that AIM can be extended to support indoor spaces with arbitrary ranges and layouts without loss of accuracy by deploying distributed microphone arrays.

## 1 INTRODUCTION

Location information are crucial for drone operation, regardless of the application and target deployment environment [5, 20, 26, 36,

---

*Yuan He is the corresponding author.

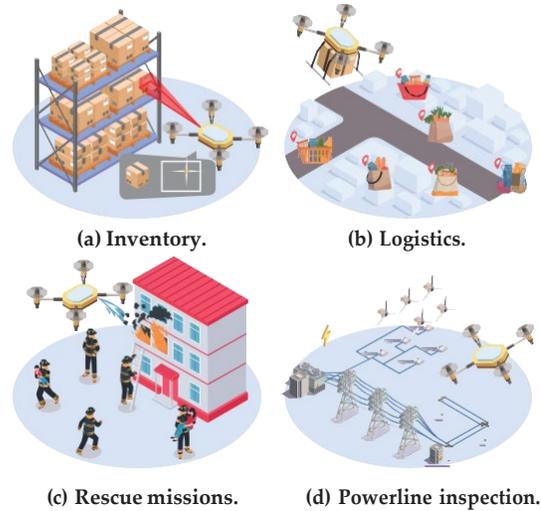

(a) Inventory.　　(b) Logistics.

(c) Rescue missions.　　(d) Powerline inspection.

**Figure 1: Example applications of drones.**

54, 56, 58]. For example, in Fig. 1, a drone for cargo inventory needs location information to determine the position of the cargo relative to its own. When performing drone deliveries, a drone must follow the predefined route and land at the right target location for the drop-off. In rescue missions, a drone needs location information to operate most efficiently during the intervention. To inspect the powerline, a precise drone location is needed to report where the anomaly is.

Location information must be *accurate*. This requirement is essential: errors in location estimates may not just degrade system performance, but represent a safety hazard as the drone's own movements are largely determined by location information [10].

**The indoor challenge.** In outdoor settings, GPS is arguably mainstream [21]. The indoor setting, however, represents a completely different ballgame, as discussed in Sec. 2.

Radar-based approaches [19, 45], for example, work both indoors and outdoors. Their spatial resolution is limited so that it is generally difficult to localize small-size drones. Further, objects in the target environment easily interfere with the radar signals [15], degrading the accuracy. RF-based localization approaches [4, 39] require installing wireless transceivers on the drone and reengineering the flight controller. Inertial measurement methods [24, 28, 35] are useful when absolute localization is unavailable, but the accumulation of errors likely becomes an issue. Infrared-based systems

Yimiao Sun[1], Weiguo Wang[1], Luca Mottola[2], Ruijin Wang[3], Yuan He[1]

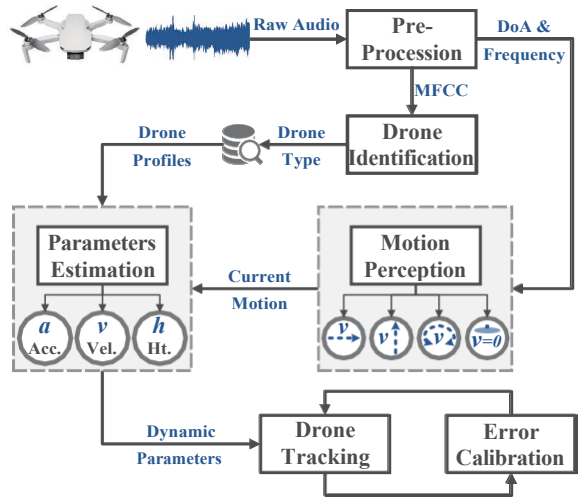

Figure 2: AIM workflow.

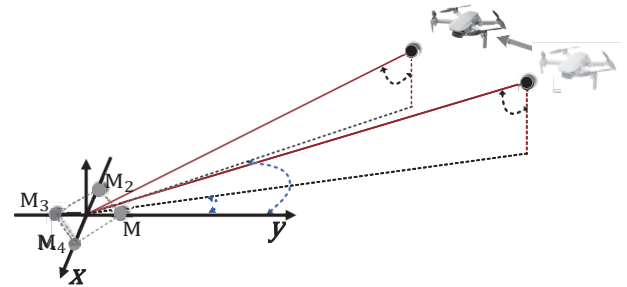

Figure 3: Schematic diagram of AIM in action.

require dedicated hardware on the drones and corresponding software changes on both drones and control stations [7].

In contrast to existing literature, we present Acoustic Inertial Measurement (AIM), a technique to localize and track drones using the acoustic signals naturally produced by the drone propellers [2]. AIM is *entirely passive*: it requires *no* additional hardware and *no* software changes on the drones. Due to the features of acoustic signals, AIM works *also in NLoS scenarios* with much better performance than the few existing systems that would be inapplicable in these settings. Further, AIM works with a single microphone array but may be extended with ease to support spaces with arbitrary ranges and layouts by deploying distributed arrays.

To achieve accurate localization and tracking of drones in complex indoor settings, without requiring the drones to generate any type of signals deliberately, we must tackle key challenges:

1) A single microphone array can only acquire one direction of arrival (DoA), which denotes the drone's direction relative to the array; this information alone is insufficient for location calculation.

2) The only input to AIM is the propellers' sound of the drone; how to infer the drone's location and motion from this single acoustic signal is an open problem.

3) In complex indoor environments, the acoustic channel between the drone and the microphone array is easily interfered by ambient noise and obstacles, or travels along NLoS paths.

**AIM.** We address these issues based on the fundamental observation that the rotating propellers create a *dual acoustic channel*: from the microphone array's view, the propellers are regarded as the sound source, so the DoA of sound denotes the *orientation* of the drone. At the same time, the propellers are also high-speed rotating machinery, so the frequency properties of the sound actually correspond to the rotating state of the propellers, which in turn determines the drone's *motion*. Obtaining *orientation and motion* information allows us to track the drone's location continuously. We further articulate the features of a drone's sound signal in Sec. 3.

Fig. 2 illustrates AIM's workflow and serves as a road-map through the rest of the paper. Consider for example the situation shown in Fig. 3, where a drone flies from $S_t$ to $S_{t+1}$. A single 4-microphone array with elements $M_1 \ldots M_4$ is deployed to capture the acoustic signals naturally produced by the drone during the flight. The raw signal is first pre-processed to extract the characteristics of the acoustic signal, for example, DoA, frequencies, and Mel-Frequency Cepstral Coefficients (MFCC). As further illustrated in Sec. 4, DoA and frequencies help deduce the drone's current motion, whereas MFCC is utilized for identifying the specific drone structure, for example, a quadcopter as opposed to an octocopter, and then loading the corresponding profile information (e.g., mass) from a database. By feeding the drone's profiles into a set of dynamic equations we formulate, we estimate its dynamic parameters, that is, acceleration and velocity, as described in Sec. 5. The drone's location is calculated consequently. To reduce error, we adopt a dedicated Kalman filter and the Interquartile Range rule (IQR), also described in Sec. 5.

We implement the workflow of Fig. 2 using off-the-shelf microphone arrays and perform an evaluation using a commercial drone under varied settings, as reported in Sec. 6. Results demonstrate that the mean error of AIM is 46% lower than commercial UWB-based systems in complex indoor scenarios, where state-of-the-art infrared-based systems cannot even work. Sec. 7 further provides additional evidence of the performance and practical applicability of AIM by reporting insights and performance from a real-world deployment in a warehouse. We show, for example, how distributed microphone arrays allow the system to extend the operating range, work around obstacles, and operate in severe NLoS settings. This functionality is achieved essentially with *no accuracy penalty*.

We conclude by discussing practical issues of applicability and general use in Sec. 8 and with brief concluding remarks in Sec. 9.

## 2 RELATED WORK

The distinctive feature of our work is to perform drone localization and tracking using acoustic signals. We briefly survey existing efforts in either field.

**Drone localization and tracking.** GPS is a mature approach widely used for drone localization and offers meter-level localization accuracy, but its application indoors is extremely difficult [16].

RF signals are explored for drone localization [34, 39], with average errors over 10 m. Methods based on optics [7], UWB radios [8] and vision [43, 49] can be applied for both indoor and outdoor drone localization, achieving more accurate results. However, methods based on optics and vision vastly assume line-of-sight (LoS) conditions and are sensitive to lighting conditions. UWB radios may partly operate in NLoS, yet their performance vastly degrades in the presence of equipment that absorbs or scatters UWB signal. [15].

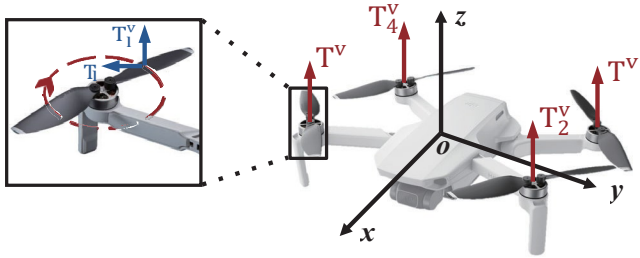

Figure 4: Quadcopter drone structure.

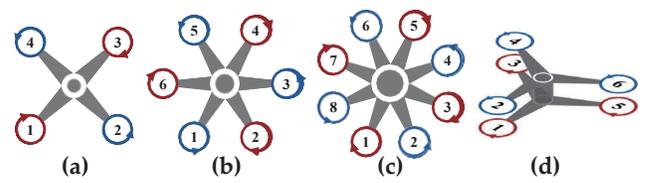

Figure 5: Typical structures of four drone types: (a) quadcopter; (b) hexacopter; (c) octocopter; (d) Y6. Different colors represent different directions of rotation.

In contrast, AIM enjoys the fact that acoustic signals may be fruitfully employed also in NLoS settings. For example, Mao et al. [33] attach two speakers on the drone to emit Frequency-Modulated Continuous-Wave (FMCW) signals, used to estimate the distance between the drone and a mobile phone. As for AIM, it does not install any extra equipment on the drone. Other efforts [9, 14, 31] only regard the drone as a mobile sound source and deploy 3D or large microphone arrays to estimate its location. Compared with these techniques, we explore the theoretical connection between the drone's sound and its motions, deduce the drone's dynamic parameters, such as velocity and acceleration, from its sound and track the drone by using only a small 2D microphone array.

Various signals emitted by the drones may be employed to identify the type of drone. Matthan detects drones by observing the characteristics of the transmitted wireless signals [37, 38]. Bleep [2] embeds FMCW signals in the PWM signals of the drones' motors as a side channel for drone communication, allowing each drone to be identified based on a unique FMCW fingerprint. Both DronePrint [25] and SoundUAV [42] utilize data-driven approaches to identify the drone via its distinct acoustic characteristics during flight. In AIM, drone identification is not the ultimate goal, but rather a necessary step to perform localization and tracking.

**Acoustic signals in localization and tracking.** Several works demonstrate the use of acoustic signals for localization and tracking [12, 27, 50, 52]. With a single microphone array, Voloc [44] aligns the multi-path DoA estimation for accurate localization of indoor acoustic sources; Symphony [51] extends this method to localize multiple sources by leveraging the prior-known layout of the array. PACE [11] localizes multiple mobile users simultaneously by leveraging structure-borne and air-borne footstep impact sounds. These works assume that the localization target and the microphone array are on the same plane or that the target's altitude is known, to solve a bi-dimensional localization problem. Differently, we exploit the signal feature in both the spatial and frequency domains, achieving *three-dimensional* localization with a single array.

Recent works adopt wearable devices for tracking, such as smartwatches and earphones. SoM [57] tracks the wrist using a smartwatch with IMUs and employs the smartphone to send beacons for error calibration. Ear-AR [55] uses the IMU in earphones and smartphones to track the indoor user's location and gazing orientation. When the embedded microphone and speaker in the wired or wireless earphones have already formed a transceiver pair, EarphoneTrack [13] proposes to track either the microphone or speaker with this pair. Unlike what we do with AIM, these approaches are effective only in the short range, specifically between wearable devices and users' smartphones.

## 3 THE SOUND OF DRONES

In this section, we explore the features of a drone's sound signals and how they relate to motion.

### 3.1 Key Features

Drone propellers are designed to displace the air around them. The resulting pressure gradient creates a force vector. We model the connection between the sound of the drone's propellers and its physical structure.

Fig. 4 illustrates the most common drone structure, that is, the one of a quadcopter composed of two orthogonal arms. A propeller is mounted at either end of each arm. The force vector obtained by the propeller rotation can be decomposed into a vertical component $T_i^v$ and a horizontal component $T_i^h$.

The vertical component lifts the drone and can be calculated as $T_i^v = k^v f_i^2$, where $f_i$ is the rotation frequency of the $i^{th}$ propeller and $k^v$ is a constant related to the lift coefficient. The drag force $T_i^h$ horizontally controls the rotation of the body and can be calculated as $T_i^h = k^h f_i^2$, where $k^h$ is a constant related to drag coefficient [29, 32]. The lift forces of all propellers follow the same direction, while the drag forces of adjacent propellers are opposite to compensate for the torque otherwise generated, which induces spinning.

The sound produced by the propellers is highly correlated with the frequency $f_i$ of each motor. Because each propeller has multiple blades, two in most cases, the fundamental frequency of the sound is not the rotation frequency $f_i$, but the blade passing frequency (BPF). The BPF is defined as $f_i^{BPF} = nf_i$, where $n$ is the number of blades. In addition to the BPF, harmonic frequencies may also be observed as an integer multiple of the BPF [6, 22].

If we can capture the drone's sound and obtain the BPF as well as its harmonics, we may then estimate the rotation frequencies $f_i$, and thus the forces exerted by each propeller. Using a model of the drone's physical dynamics, which is necessarily a function of its mechanical structure, we may also estimate its direction and motion. This is the essence of the frequency-based localization and tracking in AIM.

Yimiao Sun[1], Weiguo Wang[1], Luca Mottola[2], Ruijin Wang[3], Yuan He[1]

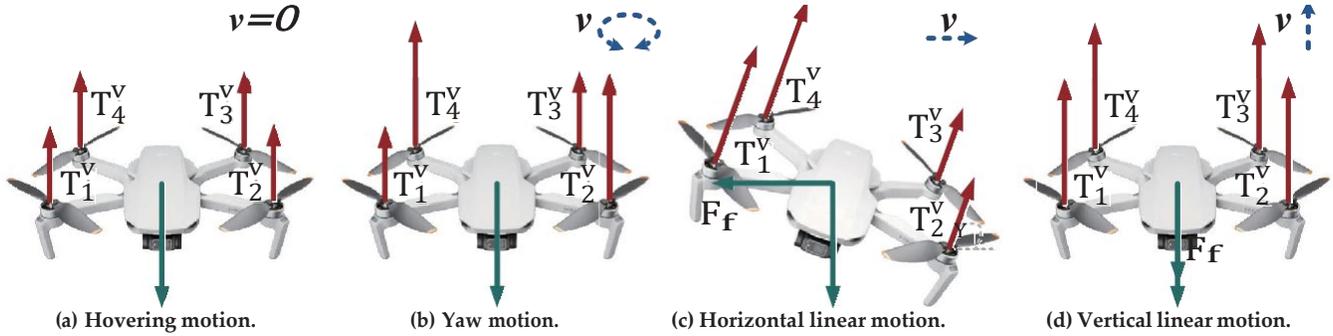

**Figure 6: Force analysis of basic drone motions.**

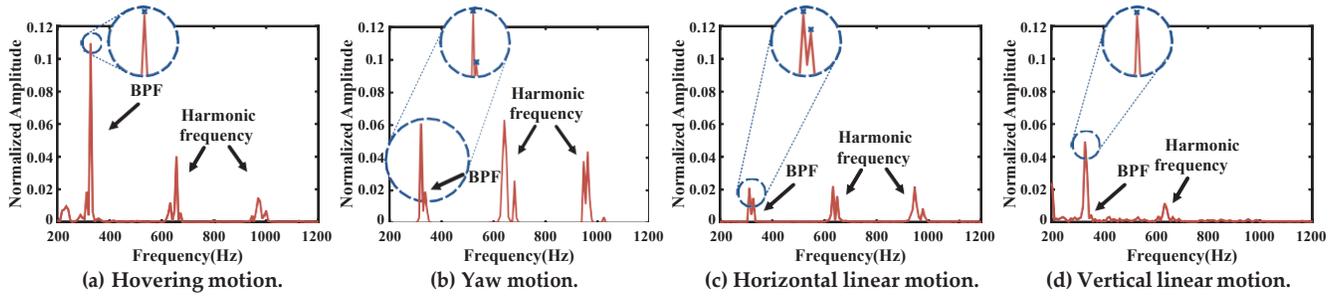

**Figure 7: Acoustic spectrum of basic drone motions.**

## 3.2 Sound and Motion

We analyze here the inner relationship between the drone's sound and its physical motion.

We theoretically analyze the acoustic properties of four common drone structures, shown in Fig. 5. Drone flights are composed of four basic motions: hovering, yaw, horizontal linear motion and vertical linear motion, as depicted in Fig. 6. Interestingly, we find that these basic motions exhibit different acoustic properties in the frequency domain because they are performed by changing each motor's rotation frequency $f_i$ differently. In the following, $N = 4$, 6 or 8 depending on the drone structure among the ones in Fig. 5.

**Hovering:** in the absence of environmental effects requiring compensation, all propellers rotate at the same frequency to maintain the vertical and horizontal balance, so the drone remains stationary. Therefore, we have $f_i = f_j, 1 \leq i, j \leq N$.

**Yaw:** propellers operate in pairs, shown by different colors in Fig. 5. Each pair rotates at the same frequency, creating a rotational momentum while maintaining the vertical balance, which makes the drone rotate around the center. Thus, we have $f_{2i-1} = f_{2j-1} \neq f_{2i} = f_{2j}, 1 \leq i, j \leq \frac{N}{2}$.

**Horizontal motion:** propellers operate in pairs again, this time to tilt the body while maintaining the vertical balance. Then the drone moves horizontally. We use parentheses to indicate equal frequencies for brevity. When the drone tilts forwards or backwards, that is, it pitches, we have $(f_1 f_2)$ $(f_3 f_4)$ for quadcopters, $(f_1 f_2)$ $(f_3 f_6)$ $(f_4 f_5)$ for hexacopters, $(f_1 f_2)$ $(f_3 f_8)$ $(f_4 f_7)$ $(f_5 f_6)$ for octocopters and $(f_3 f_4 f_5 f_6)$ $(f_1 f_2)$ for Y6 structures. Symmetric observations apply when the drone tilts leftwards or rightwards, that is, it rolls.

**Vertical motion:** all propellers rotate at the same speed to make the resulting thrusts greater or less than the force of gravity on the drone. Accordingly, the drone moves upwards or downwards, so we have $f_i = f_j, 1 \leq i, j \leq N$.

In the following, we illustrate how these observations may be a stepping stone to achieving accurate drone localization and tracking.

## 4 MOTION DETECTION AND DRONE IDENTIFICATION

We use the features of the sound signal in the frequency, spatial, and time domains to estimate the drone's motion and identify its structure. These two components are the basis of our system.

### 4.1 Drone Motion Detection

Based on the analysis of Sec. 3, we conduct a proof-of-concept experiment to check whether the four basic motions can be distinguished by the sound characteristics. In this experiment, we use a DJI Mini 2 quadcopter and a microphone to receive the acoustic signal.

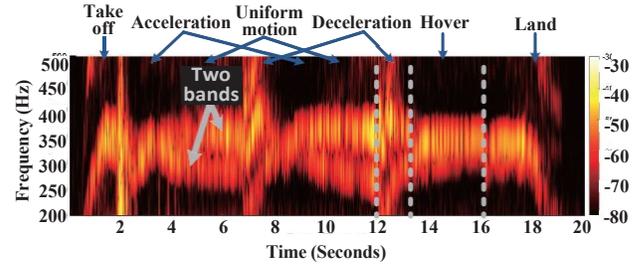

**Figure 8: The full spectrogram of the sound during a complete flight.**

Fig. 7 shows the spectrum of the acoustic signal corresponding to the motions of Fig. 6 and conforms to our understanding of the drone's dynamics. Specifically, we observe two peak fundamental frequencies in the case of yaw and horizontal motion. In comparison, there is only one peak fundamental frequency in the case of hovering and vertical motions.

We show the spectrogram of the acoustic signal of a complete flight in Fig. 8, as the drone sequentially performs different kinds of motions. The single frequency band is rising up or down when the drone takes off or lands, that is, in the case of vertical motion. In contrast, the band is split when the drone performs horizontal motion, including acceleration, uniform motion, and deceleration. For hovering, a single band is present on the spectrogram. These observations are consistent with our previous analysis.

Exclusively based on frequency domains, we can only classify the four motions into two categories, depending on the number of peak fundamental frequencies. To resolve this ambiguity, we further leverage the spatial information of the sound. Crucially, we note that the drone spatial coordinates are stable during hovering or yaw, while they change during vertical or horizontal motion. The change in position may be detected by the sound's DoA, as further elaborated in Sec. 5.1. By combining the information obtained from the number of peak fundamental frequencies and DoA as shown in Tab. 1, AIM can correctly discern the four basic motions.

Detecting the four basic motions is vastly sufficient to localize and track drones in a multitude of indoor drone applications, including most of those we mention in the Introduction. In indoor settings, for example, warehouses or smart factories, planning of robot movements—not just drones—is most often achieved by sequentially combining the four basic motions. This is beneficial in at least two respects: *i)* it matches the regular physical layout of the target deployment scenarios; in a warehouse, for example, shelves are side-by-side horizontally laid and goods are stacked vertically [18, 53]; and *ii)* it greatly simplifies path planning, yielding much more scalable systems [30].

To further improve the accuracy in detecting the four basic drone motions, we further observe that high-frequency harmonics share similar characteristics with the fundamental frequencies. Because the noise in the low-frequency band is usually stronger than that in the high-frequency band, the harmonics may experience less noise than the original BPF. Thus, we estimate the BPF from the weighted average of both the fundamental frequencies and the harmonics, which are weighted by their amplitudes. For hovering, a single band is present on the spectrogram.

**Table 1: Classification scheme of the four motions.**

|  | **Single**-Peak | **Multiple**-Peak |
|---|---|---|
| **Unstable** DoA | Vertical linear motion | Horizontal linear motion |
| **Stable** DoA | Hovering motion | Yaw motion |

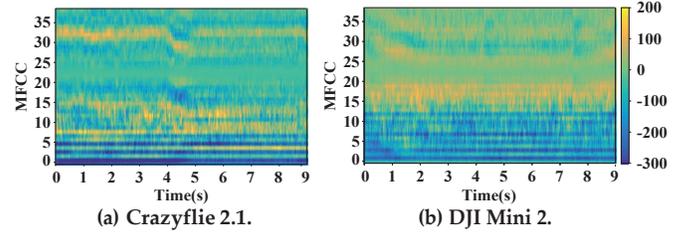

(a) Crazyflie 2.1.      (b) DJI Mini 2.

**Figure 9: MFCC of different drones.**

## 4.2 Drone Structure Identification

Several types of drones may be employed to perform different functions simultaneously, for example, assigning drones with large load capacity to carry cargoes and drones with long battery life to monitor the environment. Generally, every type of drone is assigned a specific task and a predefined route. AIM must identify the specific drone structure correctly once it is captured by the microphone array.

We observe that the sound characteristics of a drone can be represented by Mel-Frequency Cepstral Coefficients (MFCC) in the frequency and time domain. We can use this information to distinguish one drone from the others. For example, Fig. 9 shows the MFCC features of two different drones, whose energy distributions are different among MFCC vectors. Thus, AIM extracts the MFCC of a drone's sound and borrows the method proposed by Kolamunna et al. [25] to train a Long Short-Term Memory (LSTM) neural network for drone identification.

The profiles of drones serving a warehouse are pre-stored in the database. Once the drone is identified, the corresponding profile will be fed to dynamics equations for position estimation, which will be introduced next.

## 5 DRONE TRAJECTORY TRACKING

We articulate here how to combine information from the drone dynamics with the input from acoustic signals to achieve accurate drone localization and tracking. We further illustrate our system's operation in NLoS settings and how we use a dedicated Kalman filter to tame tracking errors.

### 5.1 Tracking Model

We first derive a dynamic drone model, which we use as a basis for tracking. We consider a quadcopter as an example for intuitive analysis, but the analytical process would be exactly the same for other drone structures.

**Yaw.** In this case, $(T_1^h + T_3^h) - (T_2^h + T_4^h) \neq 0$, which causes the rotation of the fuselage, as shown in Fig. 6(b), and two BPF peaks. During the rotation process, the moment of inertia $I$ reflects the magnitude of inertia and is regarded as a constant. We can thus obtain the angular acceleration $\beta_t$ at time $t$ by solving the equation:

$$\frac{k^h}{n^2} \left( \sum_{i=1}^{N/2} (f_{2i-1}^{BPF})^2 - \sum_{i=1}^{N/2} (f_2^{BPF})^2 \right) = I\beta_t \qquad (1)$$

Yimiao Sun[1], Weiguo Wang[1], Luca Mottola[2], Ruijin Wang[3], Yuan He[1]

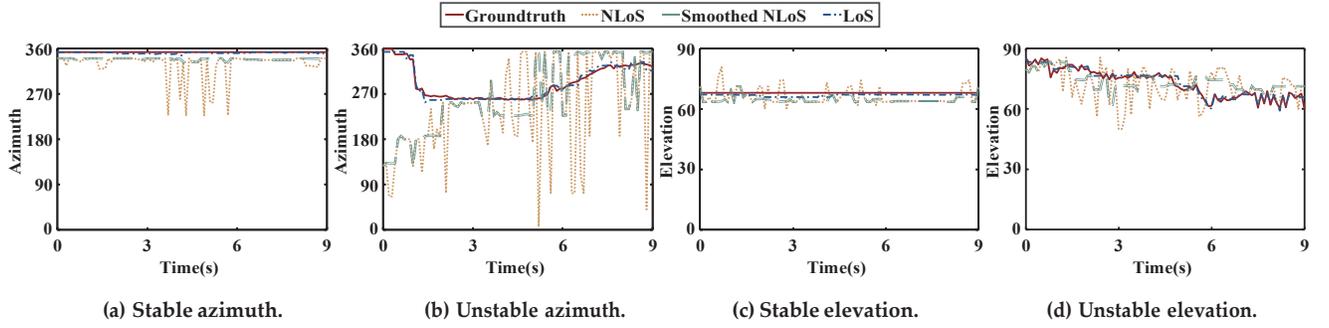

Figure 10: DoA estimation results in LoS and NLoS.

(a) Stable azimuth.  (b) Unstable azimuth.  (c) Stable elevation.  (d) Unstable elevation.

Thus, in a known time interval $\tau$, the rotation angle $\Delta\psi = \int_{\tau} \beta_t t \, dt$. However, as mentioned in Sec. 4.1, ambiguity exists if we only rely on the frequency characteristics. To solve this ambiguity, we regard the drone as a mobile sound source and leverage the microphone array to obtain spatial information. Due to the limited resolution of commercial microphone arrays, the drone is always in the far-field [51], so that we can hardly obtain accurate location information but only a DoA, including azimuth $\alpha$ and elevation $\phi$. Even in this case, DoA information is sufficient for AIM to function. For instance, DoA information captured by a uniform 4-microphone array in a squared configuration is

$$\begin{cases} \tan\alpha = \dfrac{\tau_{42}^*}{\tau_{31}^*} \\ \sin\phi = \dfrac{c}{d}\sqrt{\tau_{42}^{*2} + \tau_{31}^{*2}} \end{cases} \quad (2)$$

where $c$ is the sound velocity and $\tau_i{}^*_j$ is the time delay between microphones $M_i$ and $M_j$. We calculate the latter with the GCC-PHAT algorithm [23].

**Horizontal motion.** The rotation frequencies of two motors on the same side increase simultaneously to generate a lift force, for example $T_1^v$ and $T_4^v$ in Fig. 6(c), so that the sound contains two groups of BPF peaks, $f_1^{BPF} = f_4^{BPF}$ and $f_2^{BPF} = f_3^{BPF}$. Then the drone tilts with an angle $\gamma$, as shown in Fig. 6(c), so that we can decompose $T_i^v$ into vertical and horizontal directions. The vertical component of $T_i^v$ is balanced with the drone's gravity, so we can solve $\gamma$ with the knowledge of the drone's mass $m$ and the acceleration of gravity $g$, which are known. The horizontal component of $T_i^v$ works against the resistance $F_f = \lambda^h (v_t^h)^2$ to make the drone move horizontally, where $\lambda^h$ can be regarded as a constant related to $\gamma$. We solve the horizontal velocity $v_t^h$ and acceleration $a_t^h$ at time $t$ with the $\gamma$ by the following dynamics equations:

$$\begin{cases} \dfrac{k^v}{n^2} \sum_{i=1}^{2N} (f_i^{BPF})^2 \sin\gamma = mg \\ \dfrac{k^v}{n} \sum_{i=1}^{2N} (f_i^{BPF})^2 \cos\gamma - \lambda^h (v_t^h)^2 = ma_t^h \end{cases} \quad (3)$$

**Vertical motion.** Consider the case of climbing as an example: $f_i$, $i = 1, 2, 3, 4$ increase simultaneously to work against the gravity and downward resistance $F_f = \lambda^v (v_t^v)^2$, where $\lambda^v$ can be regarded as a constant, illustrated in Fig. 6(d). Thus, only one BPF peak is captured. Vertical velocity $v_t^v$ and acceleration $a_t^v$ at time $t$ can be determined by solving the equation:

$$\dfrac{k^v}{n} \sum_{i=1}^{2N} (f_i^{BPF})^2 - mg - \lambda^v (v_t^v)^2 = ma_t^v$$

**Finding coordinates.** Let us return to Fig. 3. The drone's coordinates at time $t$ are $S_t (h_t \tan\phi_t \cos\alpha_t, h_t \tan\phi_t \sin\alpha_t, h_t$, where the height $h_t$ is now the only unknown quantity. Fortunately, determining $h_t$ is not difficult. For two adjacent coordinates $S_t$ and $S_{t+1}$, in the case of horizontal motion, $h_t = h_{t+1}$ so that

$$|h_{t+1} \tan\phi_{t+1} - h_t \tan\phi_t| = v_t^h \tau + \dfrac{1}{2} a_t^h \tau \quad (5)$$

where $\tau$ is a predefined interval for location updating. In the case of vertical motion, we have

$$|h_{t+1} - h_t| = v_t^v \tau + \dfrac{1}{2} a_t^v \tau^2 \quad (6)$$

We solve these equations in $h_t$ and determine the complete coordinates of the drone during the flight.

### 5.2 Tracking in NLoS

Indoor scenarios likely include objects that create NLoS settings, for example, in busy warehouses. Here, the DoA information captured by the microphone array may be deviated. For instance, the yellow dashed curves in Fig. 10 depicts the estimated DoA information in NLoS settings. The severe deviation occurs in NLoS no matter whether the drone moves. In this case, traditional triangulation with distributed microphone arrays cannot work, yet alternative indoor localization systems such as UWB- and infrared-based systems may be equally prevented from working altogether in such settings.

In contrast to the state of the art, AIM can recognize if the LoS is blocked and continue to track the drone in NLoS. Despite a few outliers, the dominated diffraction or reflection path with the highest signal energy is stable when the location of the drone is unchanged, while it is irregular when the drone moves. Thus, we employ the Interquartile Range rule (IQR) [3] to eliminate outliers and smooth the estimated DoA information in a sliding window.

When the drone is hovering or yawing, the estimated DoA is smooth, as in Fig. 10(a) and Fig. 10(c), even if the observations slightly deviate from the ground truth. Instead, the smoothed DoA information is erratic when the drone is moving, as in Fig. 10(b) and Fig. 10(d). As described in Tab. 1, we use the stability of DoA information rather than the absolute values to determine the kind of drone motion in LoS. Fig. 10 provides evidence that we can employ the same criteria for the NLoS case.

To detect the NLoS setting in the first place, AIM sets a threshold to evaluate the variance of smoothed azimuth information in a time window. If the variance is beyond the threshold, we consider the LoS to be blocked, because even if smoothed, the DoA in NLoS is still unstable, which is especially evident in azimuth estimation, as shown by the green curve in Fig. 10(b).

## 5.3 Error Calibration

We employ a dedicated Kalman filter, illustrated in Algorithm 1, to tame the inaccuracies in the estimation of orientation after yawing and in absolute localization following horizontal or vertical motion.

The drone location is described by a state vector $A_t = [x_t, y_t, z_t]^T$, with $A_0$ being initialized with the first few points at the beginning of the flight (line 1). Then processing unfolds as follows:

1) We predict the subsequent state vector $\hat{A_t}^-$, that is, the a priori state estimate, according to the state transition matrix (line 3);

2) We estimate the drone's current motion following the rules in Tab. 1 as well as the current coordinate according to the dynamic equations and identified motion (line 4 5);

3) Based on the variance of the smoothed azimuth, we identify whether the LoS exists (line 6). If not, the estimated DoA information is discarded;

4) With yaw motion, possible trajectories caused by the ambiguous orientations are tracked (line 7 _ 12) until the LoS is regained. If the LoS exists now, the current coordinates can be updated with DoA, eliminating the ambiguity (line 13.15)

5) No matter whether in LoS or NLoS, the measured coordinates are fused with $\hat{A_t}^-$ to output the optimal estimate $\hat{A_t}$, that is, the a posteriori state estimate.

We proceed with describing the prototype implementation we use to gain insights on the performance of AIM in varied settings.

## 6 EVALUATION

We report evaluation results of AIM using off-the-shelf microphone arrays and a commercial drone. We describe first the implementation and evaluation settings in Sec. 6.1. Next, our investigation of AIM performance is two-pronged: Sec. 6.2 compares our system with the state-of-the-art indoor drone tracking systems and reports on their performance under different scenarios; in Sec. 6.3, we dissect the impact on tracking accuracy of environment noise, the flight range and velocity, and the number of microphones.

Our results indicate that:

1) The mean localization error of AIM in NLoS settings, arguably most realistic for indoor drone applications, is 46% lower than a UWB-based baseline;

2) Unlike an infrared-based baseline, AIM constantly provides location updates, even in NLoS settings;

3) AIM is robust to moderate noise sources in the environment, such as someone speaking;

---

**Algorithm 1:** Error reduction using Kalman filter.

1   Initialize state vector $A_0$ with the first few points;
2   **for** $t = 1, 2, 3, \ldots$ **do**
3     Predict the next state $\hat{A_t}^-$;
4     Determine current motion $Cur\_Mot$;
5     Calculate current coordinate $S_t$ as $x_t, y_t, z_t$ using dynamic equations;
6     Calculate the variance $Var$ of the smoothed azimuth in the nearest time window;
7     **if** $Var > Threshold$ **then**
8       **if** $Cur\_Mot == yaw$ **then**
9         Cache possible orientations $\Delta\psi$;
10        Track possible trajectories further;
11       **end**
12     **end**
13     **else if** $Var \leq Threshold$ **then**
14       Update $S_t$ with the measured DoA;
15     **end**
16     Fuse new estimated coordinate in $S_t$ with predicted results in $\hat{A_t}^-$;
17     Output the optimal state $\hat{A_t}$;
18   **end**

---

4) Flight range and velocity of the drone influence AIM's performance differently, yet the absolute accuracy never degrades drastically.

## 6.1 Implementation and Settings

AIM works with any layout of bidimensional microphone array to track drones of various structures. Without loss of generality, here we consider a quadcopter and two types of microphone arrays.

**Drones and microphone arrays.** We use a DJI Mini 2 quadcopter [17], shown in Fig. 11(a). The DJI Mini 2 weighs 249 g; as such, flying the DJI Mini 2 in most countries does not require a professional drone piloting license, which makes it ideal for indoor use. Each propeller is equipped with two blades. When the drone is hovering, the sound pressure level measured at a 1 m distance is empirically determined to be around 77 dB and motors run at 164 Hz, so the BPF is around 328 Hz. By default, the DJI Mini utilizes the built-in GPS for horizontal localization and an infrared time of flight (ToF) sensor to obtain vertical altitude. However, in the indoor experimental environment we use, shown in Fig. 11(b), GPS cannot work and only the ToF sensor provides useful altitude information.

We use two types of commercial off-the-shelf microphone arrays for our AIM prototype: a Seeed Studio ReSpeaker 6-mic circular array [47] and Seeed Studio ReSpeaker 4-mic array [46], shown on the upper left of Fig. 11. The inter-distance between two single microphones is 5 cm and 6.5 cm, respectively. Each microphone array is set on a Raspberry Pi 4 Model B, using a 48 KHz sampling rate. Unless stated otherwise, the results we discuss next are obtained with the 6-mic circular microphone array.

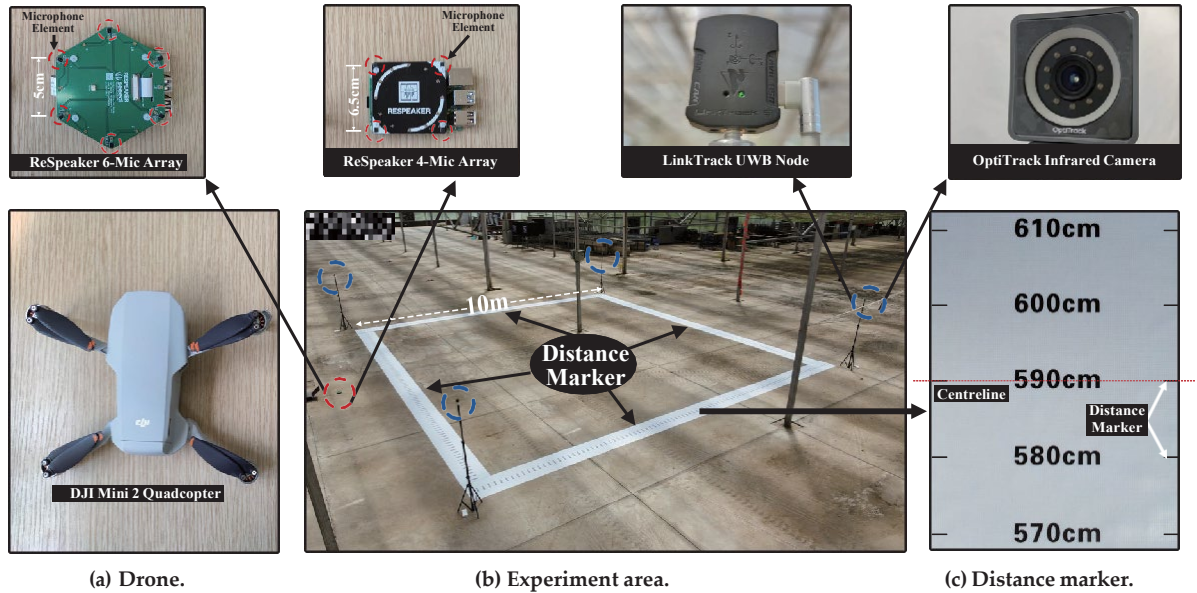

(a) Drone.  (b) Experiment area.  (c) Distance marker.

**Figure 11: Experiment settings.**

**Baselines.** To obtain ground-truth information, we take the readings of the built-in ToF sensor on the DJI Mini 2 as vertical altitude. As for the horizontal coordinates, we employ a method often used in indoor drone testbeds [1]: we lay down distance markers on the ground at intervals of 10 cm, as shown in Fig. 11(b) and Fig. 11(c). Using the downward-facing camera of the drone, we examine its view of the ground-level markers during the flight. Fig. 11(c) shows an example image captured by the drone during the experiments. Once the tick of the marker matches the centerline of the image, this reading of the corresponding maker is regarded as the real-time horizontal coordinates.

We compare AIM with LinkTrack [40], an UWB-based indoor localization system, and OptiTrack [41], an infrared-based motion tracking system, both of which are shown on the upper right of Fig. 11. LinkTrack localizes the target via triangulation. We fix a UWB tag on the drone and four UWB anchors on four tripods, then record the tracking results on a base station. OptiTrack localizes the target by converting the drone positions in bidimensional photos captured at high frequency by multiple infrared cameras to three-dimensional coordinates. We fix reflective markers on the drone and four infrared cameras on four tripods, and also record the tracking results on a base station. Whenever the drone carries a UWB tag or reflective markers, we accordingly update its tracking model and dynamic parameters.

Note that the OptiTrack system is vastly considered as state-of-the-art in indoor drone testbeds. Because of its cost, difficulty in installation, and inability to work in NLoS settings, however, it is rarely employed for real applications [1].

**Scenarios and drone mobility.** We select three scenarios for evaluation and comparison. In *Line-of-Sight* (LoS), nothing is deployed in the middle of the experiment area shown in Fig. 11(b) and every device involved in localization can establish LoS with each other and with the drone. Note how this scenario, while common in indoor drone testbeds that are in fact designed to isolate drones from their surroundings, is quite unlikely in real applications. In *Partial Line-of-Sight* (PLoS), several steel shelves stacked with various objects such as books and bricks are deployed in the middle of the experiment area. Depending on the relative position of the drone with respect to the rest of the experiment area, the LoS is blocked at times. In *None-Line-of-Sight* (NLoS), the shelves are deployed in front of every tripod hosting infrastructure node for localization. Every LoS path is thus blocked. No matter where the drone flies in the experiment field, it can not establish LoS connection to any device on any of the tripods.

We tested varied combinations of drone motions. For *horizontal motions*, we control the drone to fly along the distance maker, shown in Fig. 11(c), and keep vertical coordinates unchanged. For *vertical motions*, once the drone is hovering, we control the drone to climb or descent to a certain height, while keeping horizontal coordinates unchanged.

## 6.2 General Performance

The drone flies a 10 m × 10 m squared trajectory. We compare AIM with LinkTrack and OptiTrack in LoS, PLoS and NLoS scenarios.

Fig. 12 reports the performance of three systems. Fig. 12(a) indicates that in LoS scenarios, the mean error of AIM is 1.43 m while those of LinkTrack and OptiTrack are 0.37 m and 0.03 m, respectively [1]. AIM is, therefore, the least accurate system in LoS scenarios, which are, however, arguably rare in real applications.

Fig. 12(b) illustrates the performance in PLoS scenarios. Here AIM outperforms LinkTrack with a mean error of 1.89 m, which is 46% less than that of LinkTrack. The increase of error is caused by the lack of DoA calibration for AIM and by signal attenuation for LinkTrack. In that case, AIM can only calibrate the estimated location by the opportunistic clean DoA.

---

[1] Note that for OptiTrack, we note a difference between the error measured in our experiments and what is advertised by the manufacturer, which is below mm. The reason for this is that OptiTrack sometimes temporarily recognizes LEDs on the drones as the markers, affecting the measurements. We cannot turn off or cover these LEDs, as the drone would refuse to take off, raising exceptions in the control software.

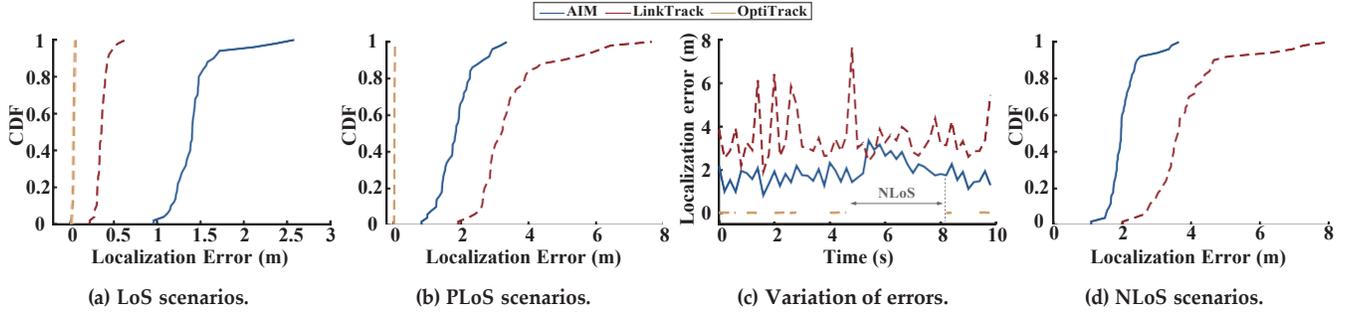

Figure 12: Performance comparison.

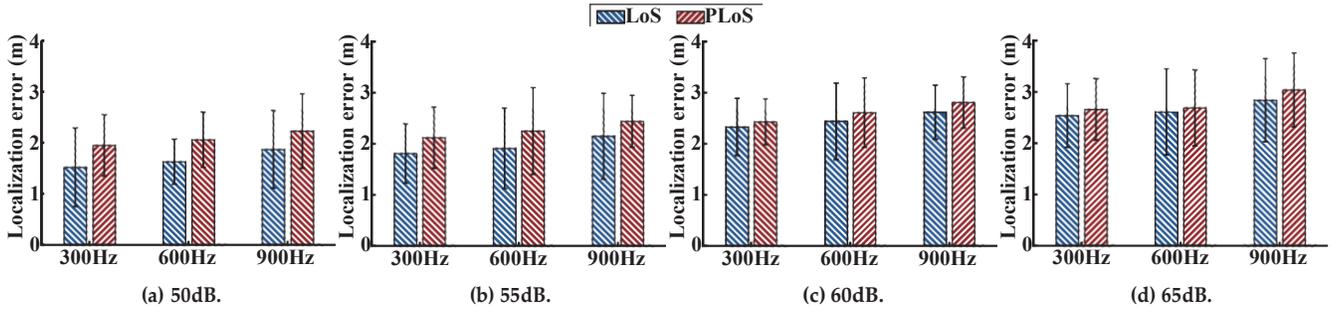

Figure 13: Impact of environment noise on accuracy.

Fig. 12(c) offers a closer view on this specific experiment by showing an accuracy comparison during a 10-sec flight, including about 2 seconds of NLoS. LinkTrack is heavily influenced by the obstacles, which absorb UWB signals. When the LoS is obstructed, OptiTrack simply does not work and produces no output. Thus, although its mean error does not increase in PLoS scenarios, OptiTrack is plainly inapplicable as completely losing the drone position even for a short among of time would be unacceptable for safe and dependable operation. Instead, the localization error of AIM suddenly increases at the beginning of the NLoS sting, but gradually decreases later, without ever losing the target.

In NLoS scenarios, shown in Fig. 12(d), we only compare AIM with LinkTrack because OptiTrack produces no output for the entire duration of the experiments, because of the aforementioned reasons. The mean error of AIM increases to 2.08 m but it is still lower than that of LinkTrack, which is almost twice as much at around 4 m.

Note how the progression through different scenarios in our discussion, from LoS in Fig. 12(a) to NLoS in Fig. 12(d), reflects increased realism in indoor drone applications. NLoS settings are indeed expected to abound when drones fly in complex physical environments. These settings are precisely where AIM reaps the greatest benefits compared to the baselines: its performance degradation, indeed, is much less pronounced compared to LinkTrack, wheres it can supply continuous location updates, unlike OptiTrack.

## 6.3 Factors Influencing Accuracy

We analyze the impact of three different factors on localization accuracy, that is, noise in the environment, the flight range and velocity, and the number of microphones.

**Environment noise.** We examine the performance of AIM in noisy conditions. We place a noise source 2 m away from the microphone array. To study different degrees of interference, we set the volume of the noise source to 50 dB, 55 dB, 60 dB and 65 dB, respectively. We broadcast noise with three different center frequencies, that is, at 300 Hz, 600 Hz and 900 Hz, to simulate interference on the BPF and its harmonic frequency.

The results in Fig. 13 indicate that, as expected, the localization accuracy degrades as the frequency of the noise or the SPL of the noise increases. This is because AIM weights the BPF and its harmonics according to their amplitude and sums them up to obtain the final frequency, which is the input of dynamic equations. In general, BPF and lower harmonics exhibit higher energy and thus are given higher weights. However, if the noise is at high frequency, peaks in this frequency band gain much higher weights. Therefore, the results are polluted.

Importantly, results show that AIM still maintains relatively stable performance under noisy conditions, which is sufficient to deal with common noise environments such as someone speaking, which generates ≈3.7 dB at 1 m distance. Further, multiple options exist to resist noise in practice. We may introduce a band-pass filter to filter out the noise band and continue tracking using the uncontaminated frequency band. AIM is also flexible in the deployment of the microphone array, in that no specific requirements must be

Yimiao Sun[1], Weiguo Wang[1], Luca Mottola[2], Ruijin Wang[3], Yuan He[1]

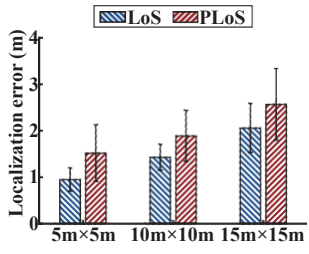

Figure 14: Flight range.

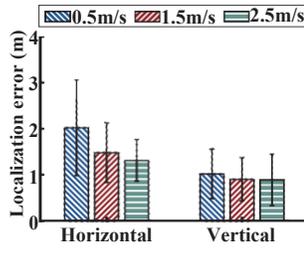

Figure 15: Flight velocity.

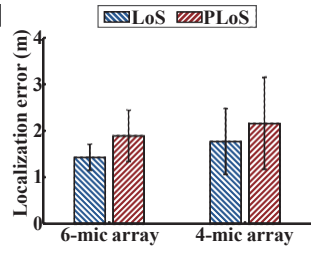

Figure 16: Number of mics.

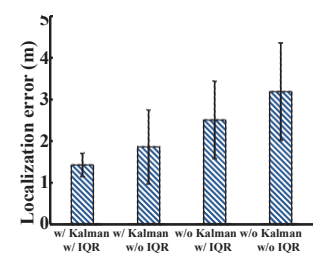

Figure 17: optimization.

fulfilled to determine where to install the array. We may simply alter its position to lessen the impact of nearby noise sources.

**Flight range and velocity.** First, we investigate the performance of AIM depending on the distance between the drone and the microphone array. We specifically test three flight paths, composed of 5 m × 5 m, 10 m × 10 m, and 15 m × 15 m square trajectories. Fig. 14 shows the results.

When the drone flies along the 5 m × 5 m square, the mean errors are 0.95 m in LoS and 1.52 m in PLoS. When the drone flies along the 10 m × 10 m square, the mean errors are 1.43 m in LoS and 1.89 m in PLoS. If the drone flies over a larger area, the signal attenuation worsens so the error increases. Thus, the results show that the mean errors in both LoS and PLoS are over 2 m as the drone flies along a 15 m × 15 m field.

Based on these results, we define 10 m as the *operational range* for the pair DJI Mini 2/ReSpeaker 6-mic. The operational range is an empirical value, which sets a limit on the acceptable tracking error. Note that this value may be different between different drones and microphone arrays, as it is mainly determined by the SPL of the sound produced by the drone's propellers and the sensitivity of the microphone array. The higher the drone's SPL and the array's sensitivity, the lower the tracking error in a given field and the larger the operational range.

We also perform experiments to evaluate if the drone's velocity has an impact on accuracy. The results are shown in Fig. 15. For horizontal motion, the drone's velocity influences the accuracy in that the mean error decreases as the velocity increases, while for vertical motion, the change of velocity does not significantly impact accuracy. The reason is two-fold. On the one hand, two frequency peaks must be captured for horizontal motion. Higher velocity results in larger intervals between the two frequency peaks, hence they are easier to separate out. In contrast, only one peak must be captured during vertical motion. On the other hand, every two propellers contribute to the energy of one frequency peak with horizontal motion, while all propellers generate the signal at the same frequency with vertical motion. The energy of the frequency peak in vertical motion is higher than that in horizontal motion and, therefore, results in more stable performance.

**Number of microphones.** Typical microphone arrays used for three-dimensional DoA estimation are 4-mic and 6-mic arrays. Fig. 16 compares the performance of these two types of arrays using the commercial off-the-shelf devices mentioned in Sec. 6.1.

The mean errors of the 6-mic array in LoS and PLoS are 1.43 m and 1.89 m, while those of the 4-mic array are 1.77 m and 2.16 m. Although with smaller inter-distance between adjacent microphones, the 6-mic array achieves more accurate results than the 4-mic array. The reason is that we always calculate the time difference of arrival (TDoA) between diagonal microphones rather than adjacent ones. The inter-distances of diagonal microphones of the two arrays are approximately 9.26 cm for the 6-mic array and 9.19 cm for the 4-mic array, but the 6-mic array provides higher redundancy because of the higher number of microphones. Thus, it produces lower errors.

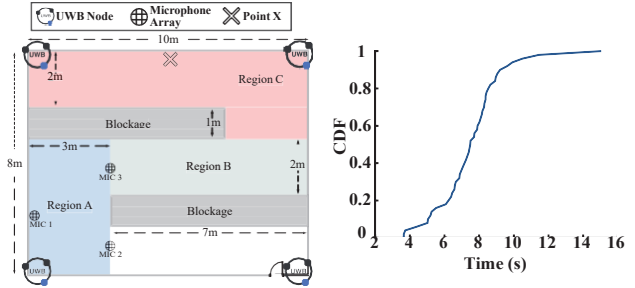

Figure 18: Layout of the ware-house. Figure 19: Time in Region C.

Note that although AIM mainly relies on the variability of DoA information rather than the accurate DoA estimation, a more accurate DoA measurement may help obtain a more accurate analysis of the DoA variability.

**Optimization algorithms.** Kalman filter and IQR method are used for error reduction in our system. Fig. 17 shows the improvement in accuracy after applying these algorithms.

If neither is used, the error of raw measurements reaches 3.19 m. If only one algorithm is employed, the error is lower when using the Kalman filter. The reason is that estimations may be wrong in some time slots without the IQR method, but they can be corrected by the Kalman filter in the long term, as most estimations are accurate. However, without the Kalman filter, the error in the whole process would not be eliminated.

Both algorithms are employed during AIM's operation, where the error reduces by 55.17% compared to raw measurements. In this setting, the mean computation delay is 34.76 ms with a standard deviation of 0.92 ms, for 50 ms signals. This latency is sufficient to update the drone's location [10].

## 7 DEPLOYMENT

We elaborate on the scalability and real-world applicability of AIM. The instrument we use to this end is a real deployment of AIM in a warehouse. The deployment is the opportunity to explore the use of multiple microphone arrays to eliminate dead zones and extend the tracking range. We also investigate the use of AIM across indoor and outdoor scenarios, assessing AIM's ability to track drones in open spaces compared to the built-in drone GPS.

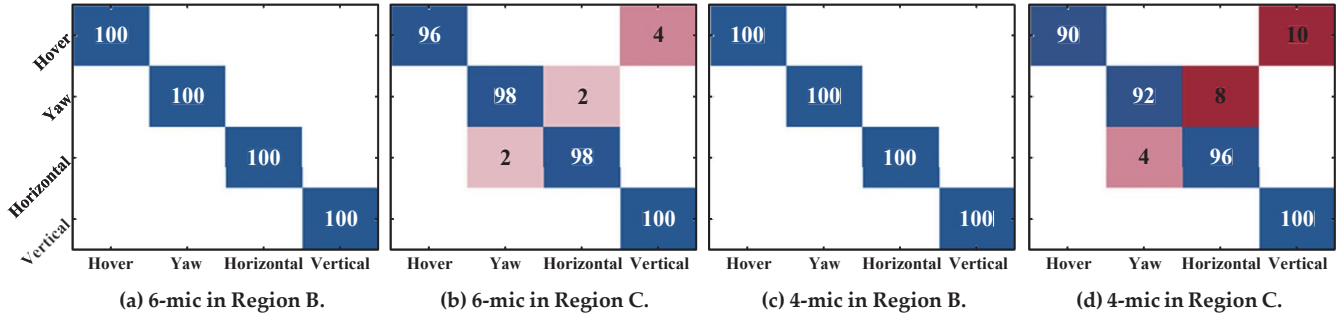

(a) 6-mic in Region B.  (b) 6-mic in Region C.  (c) 4-mic in Region B.  (d) 4-mic in Region C.

Figure 20: Motion recognition accuracy in LoS and NLoS.

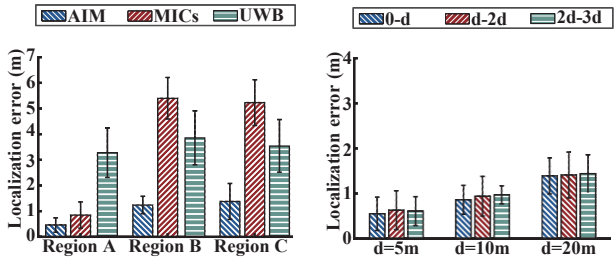

Figure 21: Accuracy in different regions.

Figure 22: Accuracy at different distances.

### 7.1 Eliminating Dead Zones

The tracking ability of a single microphone array is limited, in that it can only localize and track the drone in a small range and may introduce ambiguity when the drone is blocked in a dead zone. However, AIM can be extended easily to employ multiple microphone arrays working cooperatively.

We control the drone to fly on a 30-sec cycle to simulate working in the warehouse, whose layout is shown in Fig. 18, and Fig. 19 shows the CDF of duration when the drone is in NLoS. We deploy three microphone arrays, indicated with MIC 1, 2, and 3 to cover region A and B, leaving region C as a dead zone: the obstacle at the boundary of region C blocks the LoS between the drone and all microphone arrays. The average duration of time with the drone in NLoS conditions (i.e., Region C) is 7.44 seconds. In some situations, the duration in NLoS may be observed to be as much as half its operational time. Region C is not only a large area, but it also borders a shelf. Thus, drones operate in this region for a long time. During those times, once the drone completes a yaw, the orientation ambiguity doubles.

We compare the performance of AIM when the drone works in region B or C, depending on the type of microphone array in use. Fig. 20 shows the results. When the drone operates in region B, AIM can exactly tell which kind of motion the drone is currently performing as it operates in LoS compared to the microphone arrays. Fig. 20(a) and Fig. 20(c) demonstrate these results. The LoS is blocked in region C, thus, the drone's motions are sometimes misidentified. Although the 6-mic array is more accurate than the 4-mic array, we still hope to eliminate the adverse effects of NLoS to localize the drone more accurately. To this end, we opt to deploy another microphone array at point X in Fig. 18.

Using four microphone arrays in the deployment, we compare the localization accuracy of AIM with triangulation using the same microphone arrays and LinkTrack at three regions. Fig. 21 reports the results. In region A, triangulation achieves a fair accuracy with a mean error of 0.85 m. In comparison, AIM reports more accurate results with a mean error of 0.46 m. The reason is that AIM can fuse the results from distributed microphone arrays to output more precise and stable results. When the drone enters region B and region C, triangulation becomes inapplicable, as it returns an error above 5 m, but AIM's performance is not affected. This is because our system only requires one LoS to disambiguate or not even that, whenever the drone does not perform yaw motion in NLoS. In contrast, for triangulation to work, LoS from all microphone arrays is essentially a strict requirement.

As for LinkTrack, we set the four UWB anchors at the corners of the area to cover the whole warehouse, as shown in Fig. 18. In such a deployment configuration, LinkTrack performs poorly in all three regions because of the signal loss caused by the obstacles in the warehouse. One possible solution would be to place three or four additional UWB anchors in each region to ensure a good signal quality. However, the coordination required as the number of UWB anchors increases would require very tight time synchronization across the entire system [48], thus drastically increasing complexity. Meanwhile, the net deployment cost of this method would definitely be higher than AIM.

### 7.2 Extending Tracking Range

Some warehouses are very long and narrow. These layouts are simple and fewer dead zones likely exist. However, a single microphone array may not suffice for the whole area because of the limited pick-up range.

We explore the feasibility of deploying distributed microphone arrays in these scenarios to increase the tracking range. We line up three equally-spaced microphone arrays so that they can capture the acoustic signals of the drone once it flies in their vicinity. The inter-distance $d$ between two arrays is set to 5 m, 10 m and 20 m. This setup is instrumental to validate the viability of this technique

Yimiao Sun[1], Weiguo Wang[1], Luca Mottola[2], Ruijin Wang[3], Yuan He[1]

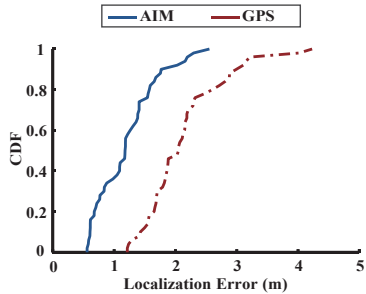

**Figure 23: Comparison with GPS.**

in general; conclusions we draw next, indeed, apply to any number of equally-spaced microphone arrays deployed linearly.

Fig. 22 shows the performance of AIM in these settings, depending on the distance $d$ between two adjacent microphone arrays as well as the flight range of the drone, shown in different colors. With a fixed $d$, the localization accuracy is relatively stable and does not decrease with the increase of flight range, which shows the feasibility of distributed microphone array. However, when $d$ increases, the tracking error increases as well, especially when $d$ jumps from 10 m to 20 m. We hinted earlier that the operational range for the pair of the drone and microphone array used in our experiments is roughly 10 m. It is then expected that the accuracy heavily decreases when the drone flies around the midpoint of adjacent microphone arrays with $d$ set to 20 m, which is two times of the operational range.

We can then conclude that $d$ should be limited within two times of the operational range of the hardware employed to ensure the accuracy at the boundary between two consecutive microphone arrays. By doing so, the tracking range can be significantly extended by deploying additional microphone arrays. Further, if the layout of the long warehouse satisfies certain conditions, distributed microphone arrays can be configured as "ZigZag", where AIM can combine the advantages of triangulation to achieve higher accuracy.

### 7.3 Transitioning Outdoor

We envision AIM can function not only indoors but outdoors. Consider the following example that simply extends the warehouse scenario. To avoid the transmission of viral diseases, some deliveries may occur on the outdoor shelves where the environment is much more ventilated. Drones may serve for inventory management indoors and perform non-contact delivery outdoors. Although the built-in GPS of the drone may work outdoors, chances are that its accuracy and stability may be barely satisfactory in this scenario, especially when location information is continuously required when rapidly transitioning from indoor to outdoor settings and the other way around. The time for the GPS to acquire a fix once the drone performs outdoor missions may significantly exceed the time available for the delivery.

We compare the outdoor bi-dimensional localization accuracy between AIM and the built-in GPS of DJI Mini 2 in an open space without any obstruction. Similar to our indoor experiments, we control the drone flying over a 10 m×10 m field and set distance markers on the ground to obtain the ground truth. Fig. 23 shows the results obtained when the built-in GPS of the drone acquires good signal quality. AIM outperforms GPS with a mean error of

1.21 m, which is 57% less than that of GPS. Given this performance of GPS in the open space, it will show even lower accuracy or can not work in a setting with obstacles and complex structures. In contrast, AIM outputs more accurate and stable tracking results no matter whether it performs outdoors or indoors.

## 8 DISCUSSION

We complete the discussion of AIM by articulating practical issues of applicability and general use.

**Operational range.** As discussed before, the operational range depends on the SPL of the drone's sound and the pick-up ability of the microphone array. However, even for small drones and microphone arrays with short operational range, AIM can continuously report the drone's location by extending the deployment of distributed microphone arrays. In practical deployments, the distance between two microphone arrays must be controlled to ensure the drone with the shortest operational range can be successfully tracked.

**Multi-drone tracking.** When multiple drones enter the same area, AIM can still track them separately if their BPF are different. Otherwise, frequency aliasing happens. We may handle this problem by borrowing ideas from existing works to discriminate different sound sources along different propagation paths [51] or to modulate the unique acoustic signature in the drone motor sound [2].

**Doppler effect.** As the drone is a mobile sound source, one may argue that the Doppler effect may represent a problem. In fact, drones cannot fly at extremely high speeds to increase lifetime and to reduce the chances of collisions [2], especially when functioning indoors. Even by assuming that the maximum speed of the drone is 5 m/s, there would be an error of less than 1.5% in the received frequency, when the sound velocity is 343 m/s. This is negligible, justifying the design choice in AIM of not compensating for the frequency shift when tracking.

## 9 CONCLUSION

We presented AIM, the first-of-its-kind passive indoor drone tracking technique that works with a single microphone array, but may also be extended to support spaces with any range and layout by deploying distributed microphone arrays. AIM innovates the acoustic tracking technique in that it fully exploits the dual acoustic channel from the drone to the microphone array, based on an in-depth understanding of the drone's dynamics and the characteristics of its acoustic signal. Through extensive experiments, we demonstrate that AIM offers strikingly better performance than state-of-the-art solutions, especially in NLoS settings, and enjoys stable performance across complex indoor environments.

## ACKNOWLEDGMENTS

We thank our anonymous shepherd and reviewers for their insightful comments. This work is partially supported by the National Science Fund of China under grant No. U21B2007 and No. 62271128, the R&D Project of Key Core Technology and Generic Technology in Shanxi Province under grant No. 2020XXX007, the Swedish Science Foundation (SSF), the Digital Futures programme (project Drone Arena), the Swedish Research Council under grant 2018-05024, and KAW project UPDATE.